\documentclass[conference]{IEEEtran}
\IEEEoverridecommandlockouts
\usepackage{cite}
\usepackage{amsmath,amssymb,amsfonts}
\usepackage{algorithmic}
\usepackage{graphicx}
\usepackage{textcomp}
\usepackage{xcolor}
\usepackage{braket}
\def\BibTeX{{\rm B\kern-.05em{\sc i\kern-.025em b}\kern-.08em
    T\kern-.1667em\lower.7ex\hbox{E}\kern-.125emX}}
\begin{document}

\title{Quantum-Enhanced Similarity Measures for Polarimetric Materials Classification}

\author{
\IEEEauthorblockN{
Sara Shojaei\IEEEauthorrefmark{2}\IEEEauthorrefmark{1},
Seyed Mohamad Ali Tousi\IEEEauthorrefmark{2}\IEEEauthorrefmark{1},
Emma Bennett\IEEEauthorrefmark{2},
Param Sangani\IEEEauthorrefmark{3},
Ali Shiri Sichani\IEEEauthorrefmark{2}, \\
Ilker Ersoy\IEEEauthorrefmark{2},
Hadi Ali-Akbarpour\IEEEauthorrefmark{3},
Filiz Bunyak\IEEEauthorrefmark{2},
G. N. DeSouza\IEEEauthorrefmark{2}
}
\IEEEauthorblockA{\IEEEauthorrefmark{2}University of Missouri--Columbia, MO, USA\\
Email: \{ssfht, STousi, enb7x6, asp9f, ersoyi, bunyak, DeSouzaG\}@missouri.edu}
\IEEEauthorblockA{\IEEEauthorrefmark{3}Saint Louis University, MO, USA; 
Email: \{param.sangani, hadi.akbarpour\}@slu.edu}\\
\IEEEauthorrefmark{1} These authors contribute equally to this work
}

\maketitle

\begin{abstract}
We present a quantum--classical hybrid pipeline for polarimetric material classification that casts this as a point-matching problem. Voxel cubes, containing polarized light reflections, are used to train an encoder to produce 32-dimensional embeddings for the voxels of the cubes. At inference, the encoder head is discarded and the embeddings are encoded as probability amplitudes of quantum states. Next, a SWAP-test circuit estimates the fidelity between each of the 32D embeddings from the query cube and a dataset of anchor cubes. The aggregated fidelity serves as  materials similarity scores, and the class of the anchor with highest aggregated fidelity is deemed as the class of the queried material. We evaluate our approach on a dataset of 23 materials ($\approx$800 samples each) derived from their Mueller matrices. The point-matching approaches from the proposed quantum SWAP-test and a classical classifier using Optimal Transport are compared. Our results demonstrate the competitive classification accuracy alongside open-set discrimination potential, establishing it as a viable path toward NISQ-based material recognition.
\end{abstract}


\section{Introduction}

Material classification is a fundamental perception task in many applications, including robotics, remote sensing, industrial inspection, etc., where accurate identification of surface material enables downstream tasks such as reasoning about physical properties, structural integrity, and safety in critical decisions.
While RGB-based approaches have advanced considerably through deep learning, they remain fundamentally limited by their reliance on color and intensity --- cues that confound material identity with illumination conditions, viewpoint,
and surface orientation.
Polarization imaging offers a physically grounded alternative: the polarimetric response of a surface, captured through its Mueller matrix, encodes reflectance properties such as dielectric constant, roughness, and subsurface scattering
that are largely invariant to illumination color yet vary distinctly across material types \cite{wolff1991,mueller_dataset}.
Despite this promise, polarimetric classifiers trained end-to-end in a supervised setting remain sensitive to distribution shift: changes in azimuth angle, illumination geometry, or sensor configuration between training and deployment reduce the accuracy, which motivates a representation that is more decoupled from acquisition conditions.

In this paper, rather than training a classical discriminative classifier on polarization images, we formulate the material recognition problem as a \textit{point-matching problem}. For that, \textit{polarization cubes}, formed by the stacking of image reflections from multiple polarizing angles, are fed to an encoder to generate $32$ dimensional embeddings for each voxel in the cube. The set of embeddings is then deemed as attributes of these point clouds, or \textit{polarization cubes}, and a new quantum-driven point-matching method is used to compute the best match between query polarization cube and a dataset of known polarization cubes (known materials). The material (anchor) in the dataset with polarization cube most similar to the polarization cube of the queried material provides the class of the same query.
This framing decouples representation learning from the matching criterion, allowing the encoder to incorporate viewpoint and illumination variation during training while the matcher operates directly on the resulting embedding geometry.

The quantum SWAP test \cite{buhrman2001,cincio2018} provides a natural primitive for estimating the similarity between query and anchors cubes. Given two quantum states $\ket{\psi}$ and $\ket{\phi}$, and their probability amplitudes computed from the embeddings in the corresponding cubes, the SWAP test estimates their fidelity using a shallow circuit: one ancilla qubit, a Hadamard gate, a controlled-SWAP, and a single measurement. Here, fidelity is the inner-product similarity $|\braket{\psi|\phi}|^2$  between quantum-encoded voxels (point clouds), making it the \textit{quantum equivalent of a soft point-matching score}.

As it will be explained in detail later, in order to reduce the number of embeddings per cube, the polarization cubes are converted to Octrees, and an OctGPT encoder \cite{octgpt2025} with a Graph Neural Network (GNN) are used to produce the embeddings. At inference, the GNN head is discarded; the embeddings at each leaf of the octree are encoded as the probability amplitude of quantum states and used in the SWAP-test algorithm above.
Finally, to evaluate our quantum approach, a classical point matching counterpart was implemented using Optimal Transport \cite{peyre2019}. This is a classical soft-matching equivalent to the proposed quantum fidelity aggregation, enabling a direct  comparison between the two approaches.

\noindent Our main contributions are:
\begin{itemize}
    \item A polarimetric cube dataset of $23$ material classes ($\approx$800 samples each), generated by varying the azimuth angle, light wavelength and polarization state over Mueller matrices from \cite{mueller_dataset};
    \item A feature extraction joined by an embedding generation pipeline producing a rich embedding matrix per material cube; trained discriminatively; and frozen at inference;
    \item A quantum SWAP-test point-matching framework that encodes voxel embeddings as quantum amplitude states, and evaluates material similarities under quantum simulation;
    \item Extensive comparison of our quantum approach against classical Optimal Transport (soft matching), Hungarian distance (hard matching), and cosine similarity baselines.
\end{itemize}

\section{Related Works}

Physics-driven polarimetric material recognition has relied on Fresnel-based metallic--dielectric separation and phase refinements at specular highlights \cite{wolff1990,chen1998}, and on Mueller-matrix decompositions (diattenuation, retardance, depolarization) and degree-of-polarization features for snapshot classification \cite{lu1996,tominaga2008}. Graphics research shifted toward acquisition, producing polarimetric BRDF formulations and dense image-based measurement pipelines \cite{kondo2020,baek2020}, while deep learning has targeted inverse rendering and supervised multimodal segmentation rather than open-world retrieval \cite{deschaintre2021,liang2022}. Robustness to unseen materials and out-of-distribution conditions remains an open gap; our work addresses it by recasting recognition as similarity-based retrieval over local embeddings rather than closed-set classification.

3D feature extraction has progressed from hierarchical neighborhood aggregators to graph and transformer backbones. PointNet++ introduced local aggregation with growing receptive fields, and O-CNN made high-resolution processing tractable with octree sparsity \cite{pointnetpp2017,ocnn2017}. DGCNN recomputed k-NN graphs in feature space via EdgeConv, and Dual Octree Graph Networks propagated graph convolutions over octree nodes \cite{dgcnn2019,dualoctree2022}. Transformers extended context further, from point-wise self-attention to discrete-VAE tokenization pretraining \cite{pointtransformer2021,pointbert2022}. OctFormer and OctGPT show that octree structure enables linear-time local-window attention and efficient multiscale tokenization \cite{octformer2023,octgpt2025}. None of these works use octree leaves as a fixed, structurally comparable descriptor set; we fill that gap with a stable 64-leaf decomposition whose embeddings are compared by optimal transport classically and by quantum fidelity at inference.

\begin{figure*}[ht]
    \centering
    \includegraphics[width=0.75\linewidth]{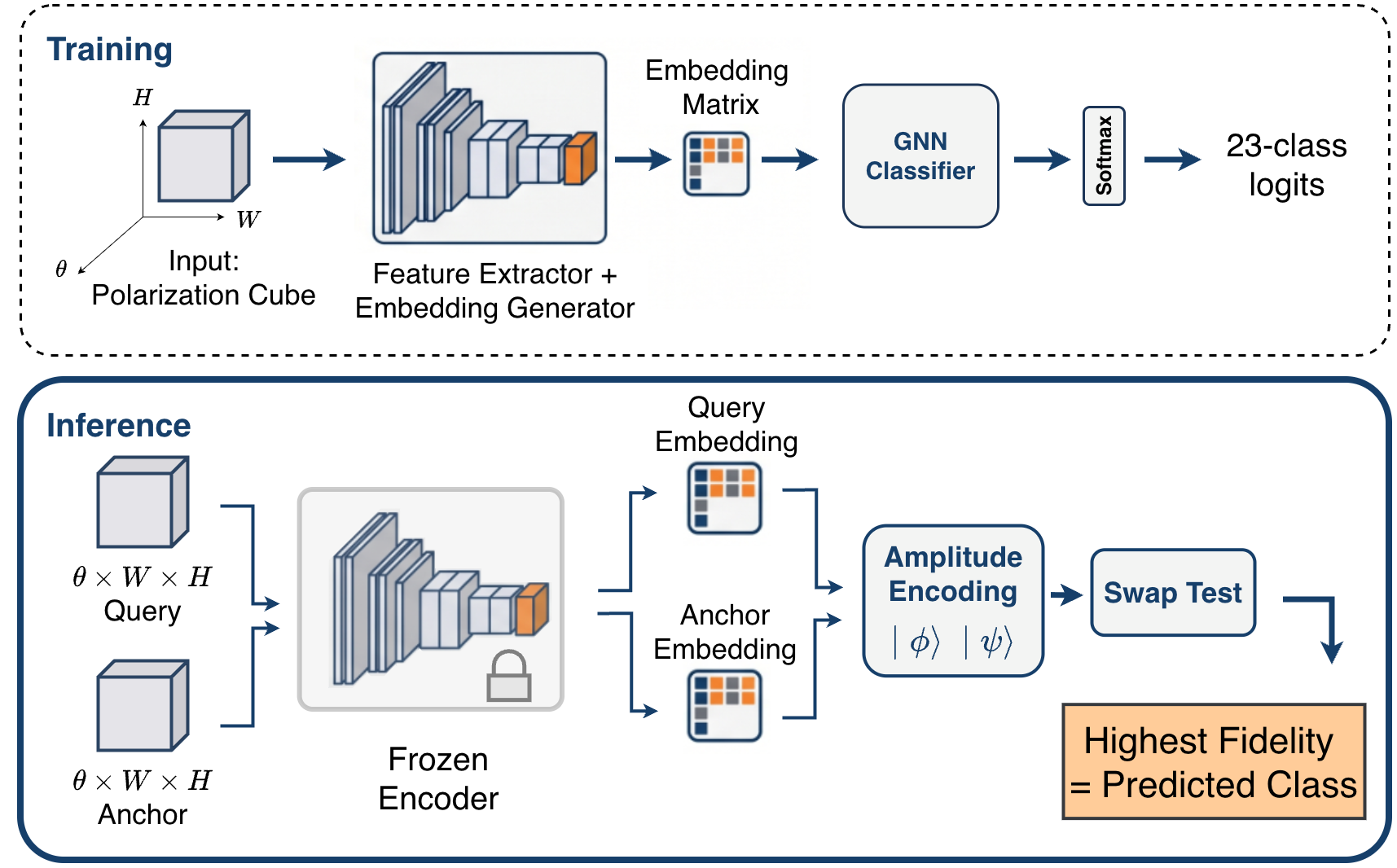}
    \caption{Overview of the proposed quantum--classical hybrid pipeline for polarimetric material classification.
    \textbf{Top (training):} A polarization cube ($\theta \times W \times H$) is passed through a feature extractor and embedding generator, producing a compact embedding matrix. A classifier head is trained end-to-end for 23-class supervised classification and discarded at inference.
    \textbf{Bottom (inference):} Query and anchor cubes are independently encoded by the frozen encoder. Their embedding matrices are amplitude-encoded as quantum states $|\psi\rangle$ and $|\phi\rangle$ and compared via the SWAP test, which estimates fidelity $F(\psi,\phi)$. The query is assigned to the class whose anchors achieve the highest average fidelity.}
    \label{fig:pipeline}
\end{figure*}

The most relevant quantum literature for our descriptor-retrieval stage is overlap-based similarity search via the SWAP test \cite{buhrman2001}, which estimates $|\langle\psi|\phi\rangle|^2$ from a single ancilla measurement. Quantum nearest-neighbor and distance-based classifiers build on this idea: Wiebe~\emph{et al.} frame retrieval around coherent inner-product estimation, Schuld~\emph{et al.} show the comparison circuit can be shallow once state preparation is available, and Basheer~\emph{et al.} apply SWAP-test fidelity scoring directly to quantum $k$-NN \cite{wiebe2014quantum,schuld2017implementing,basheer2020quantum}. Hadamard-test variants preserve the signed real part of an inner product when needed, and Euclidean-distance estimators target a different similarity metric without returning fidelity directly \cite{basheer2020quantum,schuld2017implementing,zardini2024quantum}. Practical concerns include $O(\varepsilon^{-2})$ shot cost for additive-$\varepsilon$ overlap estimation, cSWAP depth on NISQ hardware, and state-preparation overhead, motivating ancilla-free and hardware-tailored variants \cite{cincio2018,volkoff2022ancilla}. Annealing and QUBO methods for shape matching optimize geometric correspondences rather than descriptor overlap and are adjacent but not our primary lineage \cite{golyanik2020quantum,benkner2021q,bhatia2023ccuantumm}.

\section{Proposed Method}\label{sec:methodology}

\subsection{Overview}

Our pipeline has two distinct phases, illustrated in Figure~\ref{fig:pipeline}.

\textbf{Training phase:}
A polarization cube is first fed into a feature extractor and embedding generator network (top part of Figure \ref{fig:pipeline}), resulting in a set of embeddings that captures the cube's spatial and polarimetric structure at multiple resolutions.
A classifier head is then trained end-to-end for a 23-class supervised classification.
The training is discriminative and the objective is purely to learn embeddings that are compact within a class and scattered across classes, with the classification head providing the loss signal.

\textbf{Inference phase.}
Here, bottom part of Figure \ref{fig:pipeline}, the classification head is discarded and only the frozen encoder is retained. Given a query cube, the encoder produces its embedding matrix. The full matrix is encoded as the probability amplitude of a quantum state. SWAP test estimates the fidelity between the query state and each anchor state in a pre-built dataset, or library of references (seven anchors per class, computed from held-out three testing samples per class). The query is assigned to the class whose anchors achieve the highest fidelity. 

% THIS IS NOT PART OF THE PROPOSED SYSTEM/PIPELINE: Classical baselines such as Optimal Transport, Hungarian algorithm, and cosine similarity mirror this matching step, treating each material's embeddings as an empirical point distribution and computing the similarity score.
%The following subsections detail each component in turn.
Next, we present the polarization cube representation and the dataset of materials produced in this work; followed by their matrix embeddings; and finally their quantum encoding. 

\subsection{Polarimetric Cube Dataset Representation}
To construct the training dataset, we simulate the polarimetric analyzer intensity  cubes from the KAIST calibrated Mueller pBRDF measurements \cite{mueller_dataset}, covering multiple materials spanning a broad range of surface reflectance behaviours, from specular metals (e.g., chrome, brass, gold) to diffuse dielectrics (e.g., silicones, 
billiard balls, ceramics).

% \subsubsection*{Input Data}
Each material is represented by a Mueller pBRDF tensor of dimensions $361 \times 91 \times 91 \times N_\lambda \times 4 \times 4$, indexed by 
$(\phi_d, \theta_d, \theta_h, \lambda, i, j)$, with angular axes $\phi_d \in [-\pi, \pi]$, $\theta_d \in [0, \pi/2]$, and $\theta_h \in [0, \pi/2]$. Since the illumination is unpolarized, only the first column of the Mueller matrix is required, and the relevant Stokes volumes $S_0$, $S_1$, $S_2$ are extracted at load time for a chosen wavelength index. 
$S_3$ is omitted because the analyzer cube in Eq.~\eqref{eq:Stokes} is sensitive only to linear polarization; The $(S_0, S_1, S_2)$ triple therefore captures all information recoverable through a rotating linear analyzer.
Five wavelength channels are used ($\lambda$ indices $0$–$4$), and 15 light sources are sampled uniformly from the 147 available calibration lights (every 10th index). A companion calibration file provides the sphere mask, surface points, surface normals, and light source geometry.

% HA: "every 10th index": Are the lights in KAIST uniformly distributed on the sphere??

% \subsubsection*{Geometry and Incident Direction}

Surface normals are normalized and flipped from their inward-facing calibration convention ($\mathbf{n} \leftarrow -\mathbf{n}/\|\mathbf{n}\|$). 
The incident direction (from surface toward light) per pixel is computed as:
\begin{equation}
    \boldsymbol{\omega}_i = \frac{\mathbf{L} - \mathbf{P}}
    {\|\mathbf{L} - \mathbf{P}\|},
\end{equation}
where $\mathbf{L}$ is the light center and $\mathbf{P}$ is the surface point. Pixels are discarded if $\mathbf{n} \cdot \boldsymbol{\omega}_i \leq 0$ or if either direction lies below the local hemisphere ($\omega_{i,z} \leq 0$ or $\omega_{o,z} \leq 0$).
%
% \subsubsection*{Virtual Camera Directions}
% To augment the dataset beyond the single orthographic viewpoint used in the original MATLAB code ($\boldsymbol{\omega}_o = (0,0,1)$), we simulate 17 virtual camera directions parameterised by elevation from zenith $\theta_\text{el}$ and azimuth $\phi_\text{az}$:
% \begin{equation}
%     \boldsymbol{\omega}_o = \bigl(\sin\theta_\text{el}\cos\phi_\text{az},\;
%     \sin\theta_\text{el}\sin\phi_\text{az},\;
%     \cos\theta_\text{el}\bigr).
% \end{equation}
% So that 
% we construct a controlled distribution over viewpoint and illumination, and 
% the encoder is exposed to acquisition-geometry variation.
%
To expose the encoder to acquisition-geometry variation beyond the single orthographic viewpoint of the original MATLAB code ($\boldsymbol{\omega}_o = (0,0,1)$), we simulate 17 virtual camera directions parameterised by elevation $\theta_\text{el}$ and azimuth $\phi_\text{az}$:
\begin{equation}
    \boldsymbol{\omega}_o = \bigl(\sin\theta_\text{el}\cos\phi_\text{az},\;
    \sin\theta_\text{el}\sin\phi_\text{az},\;
    \cos\theta_\text{el}\bigr).
\end{equation}
% The full set of camera configurations is listed in Table~\ref{tab:camera_params}: one at $\theta_\text{el}=0^\circ$ (the default orthographic view), four at $\theta_\text{el}=15^\circ$, four at $\theta_\text{el}=30^\circ$, and eight at $\theta_\text{el}=45^\circ$.
% \subsubsection*{Local Frame and Rusinkiewicz Parameterisation}
%
A local orthonormal tangent frame $(\mathbf{t}, \mathbf{b}, \mathbf{n})$ is constructed per pixel as $\mathbf{t} = \mathbf{n} \times \hat{y} / 
\|\mathbf{n} \times \hat{y}\|$ and $\mathbf{b} = \mathbf{t} \times \mathbf{n}$, with fallback to $\hat{x}$ when $\mathbf{n}$ is nearly parallel to $\hat{y}$. 
Both $\boldsymbol{\omega}_i$ and $\boldsymbol{\omega}_o$ are projected into 
this frame. The Rusinkiewicz half-vector parameterisation is then applied:
\begin{equation}
    \mathbf{h} = \frac{\boldsymbol{\omega}_i + \boldsymbol{\omega}_o}
    {\|\boldsymbol{\omega}_i + \boldsymbol{\omega}_o\|},
    \qquad
    \theta_h = \arccos(h_z), \quad \phi_h = \text{atan2}(h_y, h_x).
\end{equation}
The incident direction is rotated by $-\phi_h$ around $z$ and $-\theta_h$ around $y$ to yield the difference angles:
\begin{equation}
    \theta_d = \arccos(\omega_{i,z}'), \qquad \phi_d = \text{atan2}(\omega_{i,y}', \omega_{i,x}'),
\end{equation}
where primes denote coordinates after the two rotations. Pixels with a  degenerate half-vector ($\|\boldsymbol{\omega}_i + \boldsymbol{\omega}_o\| 
< 10^{-8}$) are discarded.
% \subsubsection*{Mueller Lookup and Stokes Estimation}

For each valid pixel, the angular coordinates $(\phi_d, \theta_d, \theta_h)$ are mapped to tensor indices via nearest-neighbour lookup over the $361 \times 91 \times 91$ angular grid. The per-pixel Stokes parameters are:
\begin{equation}
\label{eq:Stokes}
    S_0 = M_{11}, \quad S_1 = M_{21}, \quad S_2 = M_{31}.
\end{equation}
% \subsubsection*{Analyzer Intensity Cube}
%
The intensity at analyzer angle $\beta$ is:
\begin{equation}
    I(\beta) = \frac{1}{2}\left(S_0 + S_1\cos(2\beta) + S_2\sin(2\beta)\right),
\end{equation}
evaluated at $\beta = 0^\circ, 1^\circ, \ldots, 179^\circ$, producing a 3D cube $I(\beta, y, x)$ of shape $180 \times H \times W$.
% \subsubsection*{Quality Validation}

Each generated cube is subject to four automatic quality checks before being retained: (i) at least 200 valid pixels must survive the geometry and BRDF lookup; (ii) the fraction of invalid pixels among all masked pixels must not 
exceed 95\%; (iii) the dynamic range of the cube over valid pixels must exceed $10^{-5}$; and (iv) fewer than 10\% of valid pixels may exhibit a negative mean intensity (i.e., negative $S_0$). Cubes failing any check are discarded.
%
% \subsubsection*{Dataset Scale}
The total number of generated samples per material is bounded by $N_\text{lights} \times N_\lambda \times N_\text{camera} = 15 \times 5 \times 17 = 1{,}275$ combinations, subject to the quality filters above. Across all 23 materials, this yields up to $29{,}325$ Polarization Cubes (i.e. samples) before filtering.

\subsection{Feature Extraction and Matrix Embeddings}

The feature extraction stage maps each polarimetric cube $\mathbf{I} \in \mathbb{R}^{180 \times H \times W}$ to a compact embedding matrix
$\mathbf{E} \in \mathbb{R}^{64 \times 32}$
that serves as the material representation.

\subsubsection*{OctGPT Encoder}

We re-organize the polarization cube into a hierarchical octree $\mathcal{T}$ that partitions the 3-D volume into a multi-resolution set of non-overlapping cells. At each level $d$ of the hierarchy, an occupied node $v$ contains the polarimetric signal integrated over its corresponding sub-volume. Next, an OctGPT~\cite{octgpt2025} applies a transformer-style attention mechanism across nodes, so each resulting node-assigned embedding captures both fine-grained local texture and coarse global context. The trained encoder with parameters $\theta$ produces a hidden embedding $\mathbf{h}_v \in \mathbb{R}^{d_h}$ for every node $v$ in the octree $\mathcal{T}$ of the polarization cube $\mathbf{I}$:
\begin{equation}
    \mathbf{h}_v = \bigl[\mathrm{OctGPT}_\theta(\mathbf{I})\bigr]_v.
\end{equation}

\subsubsection*{GNN Classifier Head}

The leaf embeddings $\{\mathbf{h}_v\}_{v \in \mathcal{L}}$ are treated as node features of a graph $\mathcal{G} = (\mathcal{V}, \mathcal{E})$,
where $\mathcal{V} = \mathcal{L}$ is the set of occupied leaf nodes and $\mathcal{E}$ encodes the octree adjacency structure (lateral-neighbour edges). A graph neural network $\mathrm{GNN}_\phi$ aggregates neighbourhood
information, and a linear readout head $\mathbf{W} \in \mathbb{R}^{23 \times d_h}$
produces class logits:
\begin{equation}
    \hat{y} = \mathrm{softmax}\!\left(\mathbf{W}\,
    \mathrm{GNN}_\phi\!\left(\{\mathbf{h}_v\}_{v \in \mathcal{L}}\right)\right).
\end{equation}
The full model $(\mathrm{OctGPT}_\theta, \mathrm{GNN}_\phi, \mathbf{W})$ is trained end-to-end with cross-entropy loss over all 23 material classes. The goal of this model and its training is solely to find $\theta$ and $\phi$ so that the embeddings are compact within a class and well-scattered across classes.

\subsubsection*{Inference-Time Embedding Matrix}
%\begin{figure*}
%    \centering
%    \includegraphics[width=1\linewidth]{Figures/OT_Collapse_mod.png}
%    \caption{Iterative leaf-node aggregation process applied to an octree structure (note: real octrees have 8 leaf nodes per internal node; a reduced branching factor is shown here for clarity). (a) The original tree with parent nodes (yellow), child nodes (blue), leaf nodes (green), and their associated feature vectors. (b) Sibling leaf nodes are grouped and averaged, e.g., $Avg(L_1, L_2)$, $Avg(L_3, L_4, L_5)$, and $Avg(L_6, L_7)$. (c) Each group is replaced by a single aggregated leaf, $L'_1 = Avg_{(1,2)}$, $L'_2 = Avg_{(3,4,5)}$, and $L'_3 = Avg_{(6,7)}$, reducing tree depth by one level. (d)--(e) The aggregation is applied recursively to the updated tree, merging the new leaves with remaining siblings. (f) The process, in our real octree structures, terminates when the tree is reduced to the second level of aggregated leaves.}
%    \label{fig:ot_collapse}
%\end{figure*}
At inference the GNN head and the readout weight matrix $\mathbf{W}$ are discarded; only the frozen OctGPT encoder is retained.
The leaf-level hidden states are pooled upward through $\mathcal{T}$ to its depth-2 nodes via mean aggregation.
%,as shown in a trinary-tree  in Figure \ref{fig:ot_collapse} for easy visualization.
%In detail, for each depth-2 ancestor node $k$ with descendant leaf set $\mathcal{D}_k \subseteq \mathcal{L}$:
%\begin{equation}
%    \tilde{\mathbf{e}}_k =
%    \frac{1}{|\mathcal{D}_k|}\sum_{v \in \mathcal{D}_k} \mathbf{h}_v,
%    \qquad k = 1, \dotsc, 64.
%    \label{eq:pool}
%\end{equation}
%The resulting embedding matrix
%\begin{equation}
%    \mathbf{E} = \begin{bmatrix} \mathbf{e}_1 & \cdots & \mathbf{e}_{64}
%    \end{bmatrix}^\top \in \mathbb{R}^{64 \times 32}
%    \label{eq:embed_matrix}
%\end{equation}
%is the material's complete representation.
The 64 rows correspond to the $8^2 = 64$ nodes of the collapsed octree at depth 2, each summarizing the polarimetric content of a distinct volumetric sub-region of the cube. Because the height of the collapsed octree is fixed, every sample yields an embedding matrix of identical dimensions, enabling direct pair-wise point matching between embedding matrices from different materials.
So, these matrices are flattened to 2048-dimensional vectors and $\ell_2$-normalized
for quantum amplitude encoding, as described in Section \ref{sec:swap}.

\begin{figure}
    \centering
    \includegraphics[width=0.7\linewidth]{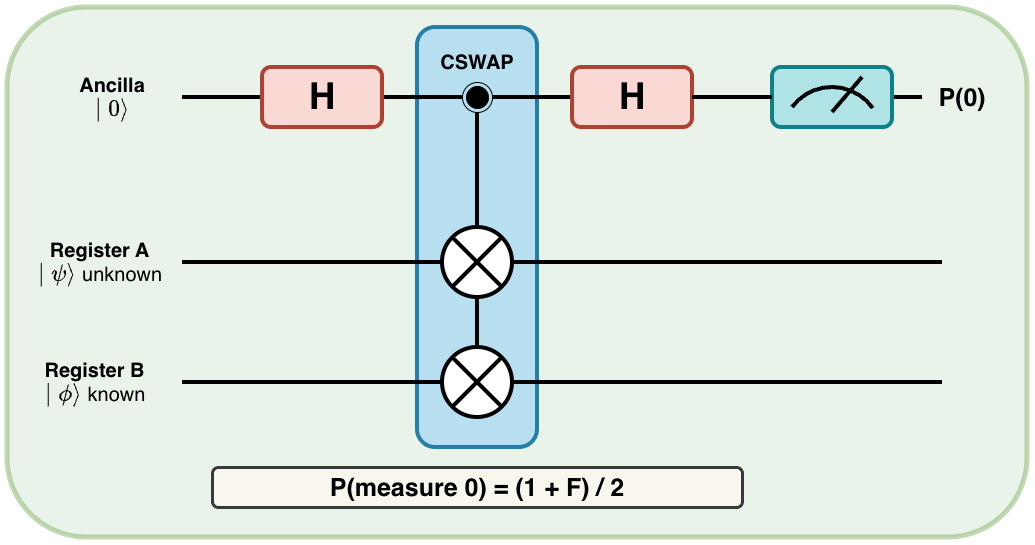}
    \caption{Quantum circuit for fidelity estimation using the SWAP test. An ancilla qubit initialized to $|0\rangle$ is placed in superposition via a Hadamard gate (H), followed by a controlled-SWAP (CSWAP) gate.
    % that conditionally exchanges the unknown state $|\psi\rangle$ (Register A) with the reference state $|\phi\rangle$ (Register B)
    After a second Hadamard and measurement, the probability of observing $|0\rangle$ is $P(\text{measure } 0) = (1 + F)/2$}
    % , where $F = |\langle\phi|\psi\rangle|^2$ is the state fidelity.
    
    \label{fig:swap-circuit}
\end{figure}

\subsection{Quantum Similarity via SWAP Test}
\label{sec:swap}

To measure similarity between two embedding matrices, now flattened as feature vectors, we compute the quantum state fidelity between their corresponding quantum encodings. Given two states $|\psi\rangle$ and $|\phi\rangle$, the fidelity is defined as:
\begin{align}
    F(\psi,\phi) = \left|\langle \psi \mid \phi \rangle\right|^2
\end{align}
\subsubsection*{Amplitude Encoding}
As mentioned above, each feature vector is derived from the depth of two node representations of 
shape $(64 \times 32)$, which is flattened to a 2048-dimensional real-valued vector $\mathbf{v} \in \mathbb{R}^{2048}$ and $\ell_2$-normalized:
\begin{equation}
    |\psi\rangle = \sum_{i=0}^{N-1} v_i |i\rangle, 
    \qquad \|\mathbf{v}\|_2 = 1,
\end{equation}
where $N = 2048 = 2^{11}$. Because $N$ is an exact power of two, no zero-padding is required. This amplitude encoding maps each normalized classical vector directly onto the computational basis amplitudes of an 11-qubit register.

\subsubsection*{SWAP Test Circuit}

In practice, the fidelity can be estimated using SWAP test, a quantum circuit that evaluates the overlap between two quantum states, as shown in Figure \ref{fig:swap-circuit}. The circuit consists of one ancilla qubit and two $n$-qubit registers (one per state), yielding a total of $1 + 2n$ qubits. With $n = 11$, the full circuit operates on 23 qubits.

The circuit proceeds as follows: (i) a Hadamard gate is applied to the ancilla; (ii) the two feature states $|\psi\rangle$ and $|\phi\rangle$ are prepared in their respective registers via amplitude encoding; (iii) controlled-SWAP (Fredkin) gates are applied between each pair of corresponding qubits, conditioned on the ancilla, and (iv) a second Hadamard is applied to the ancilla, which is then measured. The probability of measuring the ancilla in $|0\rangle$ is:
\begin{equation}
    P(0) = \frac{1 + \left|\langle \psi | \phi \rangle\right|^2}{2}.
\end{equation}
The fidelity is therefore estimated as:
\begin{equation}
    \hat{F}(\psi, \phi) = 2\,\hat{P}(0) - 1,
\end{equation}
where $\hat{P}(0)$ is the empirical frequency of the $|0\rangle$ outcome over $M = 1024$ measurement shots.
All circuits are executed under ideal conditions. The ideal (noiseless) baseline uses exact statevector simulation. 

The fidelity score provides a natural similarity measure in Hilbert space, 
capturing higher-order relationships between feature vectors that may be 
difficult to model classically.

\section{Experiments and Results}

\subsection{Implementation Details}

All experiments were conducted on the KAIST polarimetric pBRDF dataset \cite{mueller_dataset},
using the 23 material classes described in Section~\ref{sec:methodology}. The dataset comprised
19,555 polarization cube samples after filtering (approximately 
 840--876 per class), partitioned 80/20 into
15,644 training and 3,911 validation samples.
%The OctGPT encoder and GNN classification head were trained end-to-end for 23-class material
%classification on a mixture of NVIDIA GeForce RTX 3090 and NVIDIA A10 GPUs. Each input consisted of a $64\times64$ polarization cube with 180 channels corresponding to the full analyzer angle
%stack $\beta \in \{0^\circ, 1^\circ, \ldots, 179^\circ\}$, normalized per channel using
%dataset-wide z-score statistics computed over the training split. % (Figure \ref{fig:raw_cube} depicts a sample polarization cube created for green billiard material class). 

%\begin{figure}
%    \centering
%    \includegraphics[width=0.8\linewidth]{Figures/raw_cube.png}
%    \caption{A sample polarization cube created for the Green Billiard material class. Each of the 180 channels is normalized using dataset-wide z-score statistics computed over the training split.}
%    \label{fig:raw_cube}
%\end{figure}

%The octrees were constructed
%from each cube and cached after the first epoch to avoid redundant recomputation. 
Quantum circuit simulation, dataset preprocessing, and all remaining experiments were run on an Apple Mac Mini M4. At inference, the GNN head and readout matrix $\mathbf{W}$ were
discarded; only the frozen OctGPT encoder was retained. The reference library consisted of seven anchor embeddings and three query embeddings per class.
All quantum circuits were executed under ideal (noiseless) conditions using exact statevector simulation on the same Apple Mac Mini M4. Fidelity estimates were obtained from $M = 1{,}024$ measurement shots per pair, and the
query was assigned to the class achieving the highest mean fidelity across its ten anchors.

\subsection{Feature Extractor Training}

The OctGPT+GNN pipeline was trained end-to-end on a mixture of NVIDIA GeForce RTX 3090 and NVIDIA A10 GPUs for 23 material classes on the KAIST polarization dataset. Each input consisted of a $64{\times}64$ raw polarization cube with 180 channels (the full polarization angle stack), normalized per channel using dataset-wide z-score statistics computed over the training split.
Octrees were constructed from each cube and cached after the first epoch to avoid redundant recomputation in subsequent epochs.

The backbone was a VQVAE encoder (coming from the OctGPT model) producing 32-D leaf embeddings, followed by a trainable GNN classification head that operated over the octree-adjacency edges.
The full model had 729,783 trainable parameters and was optimized with Adam (initial lr $3{\times}10^{-4}$, cosine decay).

The dataset of 19,555 samples was partitioned 80/20, i.e. 15,644 training and 3,911 validation samples, with no augmentation applied to the validation set.
Training converged at epoch 51, reaching a training cross-entropy loss of 0.260 (top-1 accuracy 91.3\%) and a validation loss of 0.305 (top-1 accuracy 90.4\%), confirming that the encoder produced discriminative embeddings across all 23 material classes.
As already explained, the GNN classification head was subsequently discarded; the encoder was frozen and used solely to generate the $64{\times}32$ leaf embedding matrix that will serve as the material representation for both the quantum SWAP test and the classical baselines.

\subsection{Embedding Quality}

While the trained classifier in the first phase of the pipeline was limited to the 23 pre-trained material classes and cannot be directly applied to unseen materials, the proposed Inference phase of the pipeline was designed exactly so it can classify hitherto unknown materials. That is, by transitioning to similarity-based classification — where a query embedding matrix is matched against a library of potential matches — we should be able to achieve open-set recognition.
For that to be the case, we must guarantee that the encoder is not \emph{over-generalizing} \cite{over-generalize-1, over-generalize-2, over-generalize-3, over-generalize-4}, i.e., it must not confidently assign spurious class predictions to new (i.e. out-of-distribution) inputs as if they were familiar materials.

To verify this, we probed the frozen classifier head on two sets: (i) randomly generated Mueller matrices that lie outside the training distribution, and (ii) held-out validation samples from the 23 known classes (in-distribution).
For each sample, we recorded the maximum softmax probability (confidence) and the predictive entropy $H = -\sum_c p_c \log p_c$, normalized by $\log C$ where $C = 23$.

The results are summarized in Table \ref{tab:ood_sanity}.
Out-of-distribution inputs received a mean max-probability of 0.727 and a normalized entropy of 0.271, compared to 0.815 and 0.165 for in-distribution validation samples.
The classifier was thus \emph{less confident} and \emph{more uncertain} on unseen random inputs, which was the expected behavior of a well-calibrated encoder.
This confirms that the embedding space was not trivially collapsed, and that fidelity-based retrieval over leaf embeddings is a justified alternative to closed-set classification.

\vspace{-3mm}
\begin{table}[h]
\centering
\caption{Out-of-distribution sanity check: classifier confidence and entropy on random Mueller matrix inputs vs.\ held-out validation samples.}
\label{tab:ood_sanity}
\begin{tabular}{lcc}
\hline
& Out-of-distribution (random) & In-distribution (val) \\
\hline
Samples           & 5    & 20   \\
Mean max-prob     & $0.727$ & $0.815$ \\
Mean entropy      & $0.851$ & $0.517$ \\
Norm.\ entropy    & 0.271            & 0.165             \\
\hline
\end{tabular}
\end{table}
\vspace{-3mm}
\subsection{Classification Results}
We evaluated four point-matching strategies on the same frozen embeddings, covering three classical point-matching methods and the quantum SWAP test.
All methods performed nearest-anchor retrieval: a query's $64{\times}32$ embedding matrix was compared against every anchor in the library, and the class of the closest anchor was assigned to the query.

\textbf{Classical baselines.}
We considered three families of classical point-matching methods: \emph{\underline{Optimal Transport}}, which solves a soft assignment problem between the 64 query leaves and 64 anchor leaves, minimizing the total transport cost under the Euclidean ground metric. We evaluated Optimal Transport on $\ell_2$-normalized embeddings (\emph{OptTrans Normalized}), similar to the amplitude-based quantum state encoding, the embeddings were normalized, ensuring a fairer comparison between the quantum and classical methods;  \emph{\underline{Hungarian matching}} which solves the same assignment problem, but with a hard one-to-one constraint, using the sum of matched Euclidean distances as the score. We similarly evaluated the Hungarian algorithm on $\ell_2$-normalized embeddings; and \emph{\underline{Cosine similarity}}, which concatenates and sorts each $64{\times}32$ matrix into a 2048-D vector and computes cosine similarity. This is a strong bag-of-features baseline that ignores spatial structure of the octrees (and underlying polarization cubes). Cosine similarity also served as the closest classical equivalent to the quantum SWAP test, since the embeddings were also sorted. 

\textbf{Quantum SWAP Test.}
Each leaf embedding was $\ell_2$-normalized, sorted based on the embedding magnitude, and encoded as a quantum amplitude state.
The SWAP test circuit measured the fidelity between a query state and the corresponding anchor state to produce a per-anchor similarity score.
The circuit was simulated using the Qiskit library under noiseless statevector simulation with 1024 shots per pair.

\textbf{Results.}
Complete results are reported in Table~\ref{tab:main_results}.
The quantum SWAP test achieved 72.7\% top-1 and 90.9\% top-3 accuracy under noiseless statevector simulation, establishing quantum fidelity as a valid similarity signal for polarimetric material embeddings.
Cosine similarity, 
%which is 
the classical method structurally closest to the quantum approach, achieved 71.2\% top-1 and 93.9\% top-3, confirming that the two methods capture comparable information from the embeddings.
%geometry.
As expected, the two structure matching methods outperformed the SWAP Test, the Optimal Transport Normalized achieved 87.9\%/98.5\%, and the Hungarian Normalized reached 86.4\%/95.5\%, since they operate directly on the per-leaf set structure and solve the optimum assignment problem over the 64-leaf layout, whereas the quantum encoding compresses the sorted embedding matrix into a single amplitude state.
\begin{table}[h]
\vspace{-0.5cm}
\centering
\caption{Top-1 and top-3 nearest-anchor classification accuracy across all matching methods.}
\label{tab:main_results}
\begin{tabular}{lcc}
\hline
Method & Acc@1 & Acc@3 \\
\hline
Optimal Transport - Normalized                   & 0.879          & 0.985 \\
Hungarian - Normalized            & 0.864          & 0.955 \\
\hline
Cosine similarity             & 0.712          & 0.939 \\ \hline
\textbf{Quantum SWAP Test}               & 0.727          & 0.909 \\

\hline
\end{tabular}
\end{table}
\vspace{-0.3cm}
\section{Discussion}

\subsection{NISQ Considerations}

The amplitude-encoded SWAP test operates on $23$ qubits. While 23 logical qubits are within the nominal qubit count of current NISQ devices, the practical bottleneck is circuit depth and qubit count \textit{after transpilation}.
Each controlled-SWAP (Fredkin) gate decomposes into a sequence of two-qubit native gates; transpiling the controlled-SWAP operations across a device connectivity graph produces circuits whose depth rapidly exceeds the coherence budget of near-term hardware.
%Our results do not include a noise model precisely because the ideal state vector simulation is the relevant reference point here — any shot-based execution on real hardware would require hardware noise correction that is not yet mature for circuits of this depth.

One hardware-friendly alternative is the \emph{destructive SWAP test} \cite{volkoff2022ancilla}, which avoids the ancilla and controlled-SWAP by measuring both registers jointly after a Bell-basis rotation, recovering fidelity from coincidence statistics at shallower depth.
However, a more structurally motivated direction is to move from flat amplitude encoding toward \emph{graph-based quantum matching}.

\subsection{Limitations}

Three limitations bound the scope of the present results.
First, the dataset is derived from 23 lab-controlled Mueller matrix measurements: materials are imaged under calibrated illumination with controlled geometry, and the polarization cube samples are synthesized by varying the virtual camera viewpoint.
Performance on in-the-wild acquisitions, such as uncontrolled lighting, partial occlusion, and/or sensor noise remains unexamined.
Second, all quantum circuits are executed under ideal, noiseless statevector simulation; no noise model or error mitigation is applied.
%The accuracy gap between the SWAP test (72.7\%) and structured Optimal Transport (89.4\%) is expected: the current encoding collapses all 64 leaf embeddings into a single amplitude state, while Optimal Transport operates directly on the per-leaf set structure; a realistic hardware execution would additionally suffer from decoherence and gate error.
Third, amplitude encoding requires a state preparation subroutine whose gate complexity grows with the dimension of the input vector; at 2048 dimensions this overhead is non-trivial and dominates the total circuit cost on real hardware, a known bottleneck for amplitude-encoded quantum ML pipelines.

\subsection{Future Work}

The most immediate extension for this work should exploit the non-collapsed octree directly: each leaf embedding becoming a node feature, and matching reducing to quantum graph matching over the octree topology.
This would recover the per-leaf spatial resolution within a quantum framework and opens a natural path to quantum graph neural network architectures.
On the hardware side, the next step could be to run the destructive version of the SWAP test circuit on real IBM quantum hardware, with noise characterization across shot counts and a systematic comparison of ideal vs.\ noisy fidelity estimates.
Finally, the dataset should be expanded to cover more material classes and uncontrolled acquisition conditions, such as captured Mueller matrices from natural scenes rather than calibration setups, to evaluate robustness and generalization of both the embedding pipeline and the matching stage.

\section{Conclusion}

We have presented a quantum-classical hybrid pipeline for polarimetric material classification that reframes recognition as a point matching problem.
A polarization cube is fed to an OctGPT encoder trained with a GNN classification head to produce a $64 \times 32$ embedding matrix per sample.
At inference, the frozen encoder's output is compared against an anchor library using the quantum SWAP test.

On the 23-class KAIST polarimetric dataset, the quantum SWAP test achieves 72.7\% top-1 and 90.9\% top-3 accuracy under ideal simulation, demonstrating that quantum fidelity is a valid and meaningful similarity signal for polarimetric material embeddings.
The gap relative to structured classical methods (Optimal Transport and Hungarian, 87.9\% top-1) is expected: those methods operate directly on the per-leaf set structure, while the current quantum encoding represents the full embedding matrix as a single amplitude state, compressing spatial layout into global fidelity.
This distinction points clearly to the next step rather than to a fundamental limitation of the quantum approach.

\section*{Acknowledgment}
The quantum computation for this work was performed on the University of Missouri’s Quantum Innovation Center, in partnership with IBM Quantum and facilitated by Research Support Services at the University of Missouri, Columbia, MO. DOI: 10.32469/10355/107781. 
Other computational resources have been supported by the NSF National Research Platform, as part of GP-ENGINE (award OAC \#2322218).

\bibliographystyle{IEEEtran}
\bibliography{references}

\end{document}